\documentclass[a4paper,twoside]{article}

\usepackage{epsfig}
\usepackage{subcaption}
\usepackage{calc}
\usepackage{amssymb}
\usepackage{amstext}
\usepackage{amsmath}
\usepackage{amsthm}
\usepackage{multicol}
\usepackage{pslatex}
\usepackage{apalike}
\usepackage{algorithm2e}
\usepackage[bottom]{footmisc}

\usepackage{booktabs}
\usepackage{graphicx}
\usepackage{subcaption}
\usepackage{sidecap}
\usepackage{makecell}
\usepackage[dvipsnames]{xcolor}
\usepackage{adjustbox}
\usepackage{microtype}
\newtheorem{theorem}{Theorem}
\usepackage[pagebackref,breaklinks,colorlinks]{hyperref}
\everypar{\looseness=-1}
\usepackage[capitalize]{cleveref}
\crefname{section}{Sec.}{Secs.}
\Crefname{section}{Section}{Sections}
\Crefname{table}{Table}{Tables}
\crefname{table}{Tab.}{Tabs.}

\newcommand{\EWC}[1]{\ensuremath{\mathit{EWC#1}}}
\newcommand{\GEM}[1]{\ensuremath{\mathit{GEM#1}}}
\newcommand{\REPLAY}[1]{\ensuremath{\mathit{REPLAY#1}}}
\newcommand{\LWF}[1]{\ensuremath{\mathit{LwF#1}}}
\newcommand{\TWP}[1]{\ensuremath{\mathit{TWP#1}}}
\newcommand{\MAS}[1]{\ensuremath{\mathit{MAS#1}}}
\newcommand{\BARE}[1]{\ensuremath{\mathit{BARE#1}}}
\newcommand{\JOINT}[1]{\ensuremath{\mathit{JOINT#1}}}
\newcommand{\AOPD}[1]{\ensuremath{\mathit{AOPD#1}}}
\newcommand{\MOPD}[1]{\ensuremath{\mathit{MOPD#1}}}
\newcommand{\AAC}[1]{\ensuremath{\mathit{AA#1}}}
\newcommand{\AF}[1]{\ensuremath{\mathit{AF#1}}}
\newcommand{\OPD}[1]{\ensuremath{\mathit{OPD#1}}}

\newcommand{\changedrg}[1]{#1}
\newcommand{\changedr}[1]{#1}

\usepackage{SCITEPRESS}     

\begin{document}

\title{Benchmarking sensitivity of continual graph learning for skeleton-based action recognition}

\author{\authorname{Wei Wei\sup{1}\orcidAuthor{0000-0001-5651-8712}, Tom De Schepper\sup{1}\orcidAuthor{0000-0002-2969-3133} and Kevin Mets\sup{2}\orcidAuthor{0000-0002-4812-4841}}
\affiliation{\sup{1}University of Antwerp - imec, IDLab, Department of Computer Science, Sint-Pietersvliet 7, 2000 Antwerp, Belgium}
\affiliation{\sup{2}University of Antwerp - imec, IDLab, Faculty of Applied Engineering, Sint-Pietersvliet 7, 2000 Antwerp, Belgium}
\email{\{wei.wei, tom.deschepper, kevin.mets\}@uantwerpen.be}
}

\keywords{Continual Graph Learning, Action Recognition, Spatio-Temporal Graph, Sensitivity Analysis, Benchmark.}

\abstract{Continual learning (CL) is the research field that aims to build machine learning models that can accumulate knowledge continuously over different tasks without retraining from scratch. Previous studies have shown that pre-training graph neural networks (GNN) may lead to negative transfer\cite{hu2019strategies} after fine-tuning, a setting which is closely related to CL. Thus, we focus on studying GNN in the continual graph learning (CGL) setting. We propose the first continual graph learning benchmark for spatio-temporal graphs and use it to benchmark well-known CGL methods in this novel setting. The benchmark is based on the N-UCLA and NTU-RGB+D datasets for skeleton-based action recognition. \changedr{Beyond benchmarking for standard performance metrics}, we study the class and task-order sensitivity of CGL methods, i.e., the impact of learning order on each class/task's performance, and the architectural sensitivity of CGL methods with backbone GNN at various widths and depths. We reveal that task-order robust methods can still be class-order sensitive and observe results that contradict previous empirical observations on architectural sensitivity in CL.} 

\onecolumn \maketitle \normalsize \setcounter{footnote}{0} \vfill

\begin{figure*}[ht]
    \centering
  \begin{subfigure}[b]{0.49\textwidth}
      \centering
   
      \includegraphics[width=0.69\columnwidth]{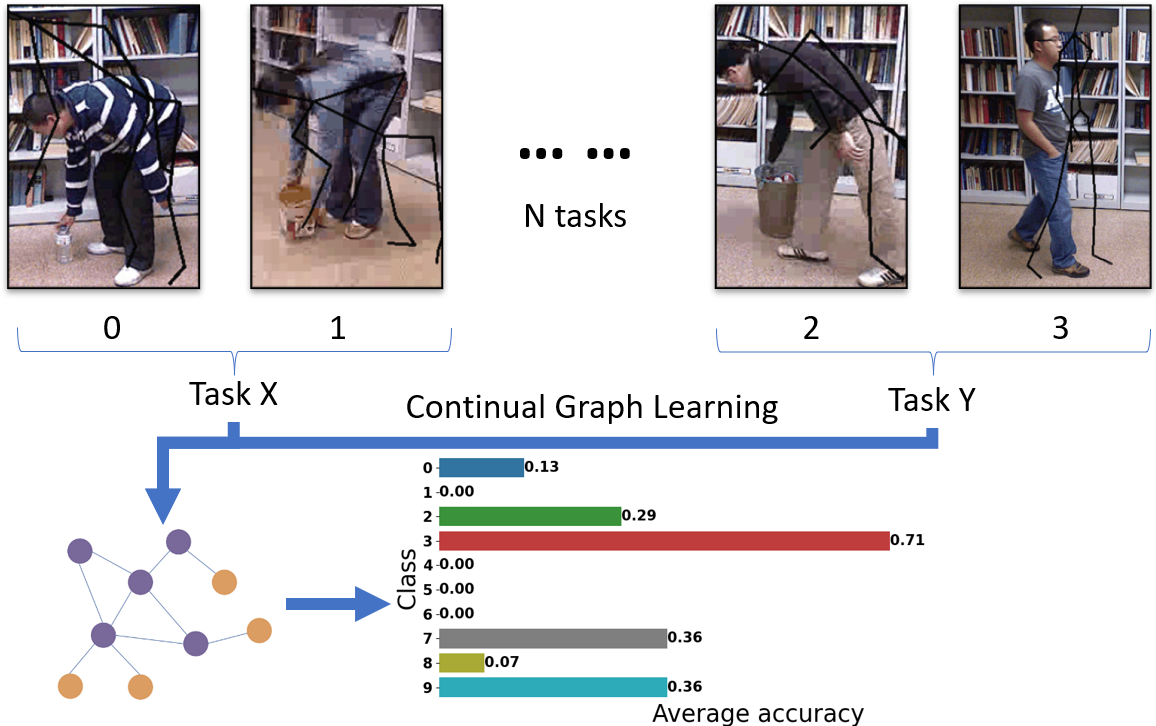} 
      \subcaption{Example of task order shuffling, tasks have same classes.}
      \label{fig:tsk_example}
  \end{subfigure}
  \begin{subfigure}[b]{0.5\textwidth}
      \centering
      \includegraphics[width=0.688\columnwidth]{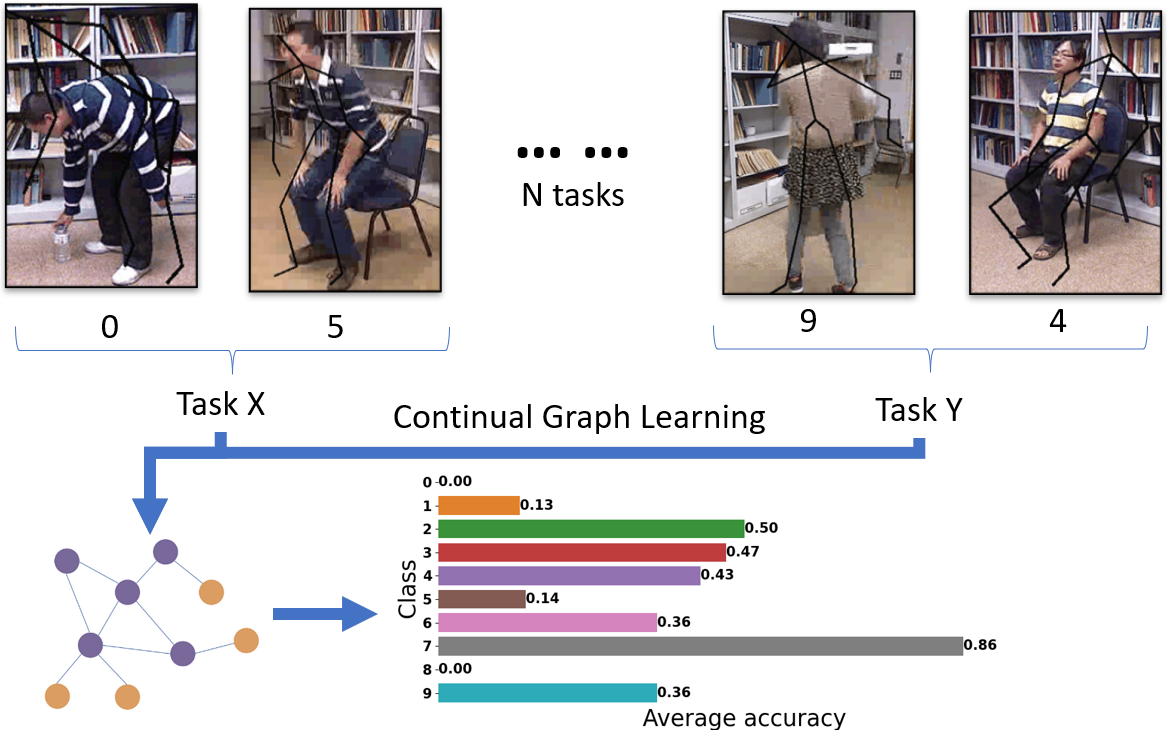}
      \subcaption{Example of class order shuffling, tasks have random classes.}
      \label{fig:cls_example}
  \end{subfigure}
    \caption{The accuracy for each class fluctuates when the task/class order for CGL changes. Classes within one task can have large accuracy differences (\cref{fig:tsk_example}, class 2/3). This is not captured by task-order sensitivity. Images from \cite{wang2014cross-ucla}.}
    \label{fig:teaser}
\end{figure*}

\section{\uppercase{Introduction}}\label{sec:introduction}

\textbf{Continual Learning (CL)} is a subfield of machine learning. It aims to build deep learning models that can continually accumulate knowledge over different tasks without needing to retrain from scratch \cite{de2021continual}. Good CL performance can not be achieved easily. The phenomenon of \textbf{catastrophic forgetting} \cite{mccloskey1989catastrophic} causes deep learning models to abruptly forget the information of past tasks, as they are no longer relevant to current optimization objectives. 

Many previous works propose new CL methods to alleviate the catastrophic forgetting phenomenon \cite{kirkpatrick2017overcoming-ewc} \cite{aljundi2018memory-mas} \cite{li2017learning-lwf} \cite{lopez2017gradient-gem} \cite{rolnick2019experience-er} \cite{isele2018selective-ser}. In comparison, the study on the sensitivity of CL performances is relatively scarce \cite{de2021continual} \cite{mirzadeh2022architecture} \cite{bell2022effect} \cite{lin2023theory}. However, from these studies, we already see a noticeable difference \changedr{in performance, measured by the average} accuracy and forgetting metrics when the backbone architecture or the learning order changes.

Research on graph neural networks (GNN) increased lately due to the availability of graph-structured data \cite{wu2020comprehensive}. However, \textbf{continual graph learning (CGL)} is still underexplored \cite{febrinanto2023graph}. Previous studies show that pre-training GNNs \changedr{is not always beneficial} and may lead to negative transfer on 
downstream tasks after fine-tuning \cite{hu2019strategies}. This is contrary to the observations in transfer learning with CNNs \cite{girshick2014rich-rcnn}. We note that training a model in a CGL setting without CGL methods is equivalent to fine-tuning the model. As GNNs show empirically worse performance on fine-tuning compared to CNNs, they may also have unique properties in CGL. This motivates our research on the sensitivity of the CL/CGL methods with graph-structured data and GNNs. Current CGL benchmarks \cite{ZhangCGLB} \cite{ko2022begin} cover many types of graphs, however, temporal graphs have not yet been benchmarked. Our benchmark uses skeleton-based action recognition datasets with spatio-temporal graphs, extending current CGL benchmarks.

In CL, the method must learn to discriminate between a growing number of classes. Often, a sequence of classification tasks with disjoint classes is used \cite{van2022three-il}. Previous literature often focuses on task-incremental learning (task-IL), where the task identifier is explicitly provided to the model during the train and test phases. We focus on class-incremental learning (class-IL), where the task identifier is not provided to the method. The method must learn to distinguish between all seen classes, i.e. the method needs to solve each task and identify which task a sample belongs to. The class-IL setting is more difficult but also more realistic, since we do not always have a task identifier associated with the collected data. 

Our work focuses on \changedr{benchmarking the performance, as well as the} order and architectural sensitivity of the CGL methods. \changedr{We measure the order sensitivity by training CGL methods on randomly shuffled task or class order and compute the corresponding metrics.} \cref{fig:tsk_example} is an example of a shuffled task order: \changedr{We pre-define a set of tasks with corresponding classes, e.g. }Task $X$ \changedr{will contain samples of} classes 0 and 1, and task $Y$ \changedr{will contain samples of} classes 2 and 3. \changedr{In contrary,} \cref{fig:cls_example} \changedr{shows} a randomly shuffled class order. Here, tasks are constructed with random classes. Task $X$ \changedr{may} consists of classes 0 and 5 \changedr{in one class order experiment, but will contain samples of other classes in another class order experiment}. We observe that the accuracy of each class differs noticeably as the learning order changes. The large difference denotes the high class-order sensitivity of the CGL method. In contrast to \changedr{standard sensitivity benchmarks}, which measures the fluctuation of the average accuracy computed across the tasks/classes. \changedr{We measure the} order sensitivity proposed by \cite{yoon2019scalable} \changedr{and extend the setting to two different scenario: task-order sensitivity and class-order sensitivity separately}. \changedr{They measure} the impact of learning order on \changedr{the performance} each specific task or class. \changedr{This can help with assessing} whether the past tasks/classes \changedr{will} have equal performance \changedr{when a CL/CGL model is deployed in real world setting, where we do not have control over the order of arrival for the tasks/classes}. It is important for the application where fairness is crucial. 

Our contributions are as follows:
\begin{itemize}
    \item To the best of our knowledge, we are the first to benchmark CGL methods for tasks involving spatio-temporal graphs, such as skeleton-based action recognition, in class-IL setting. It requires the model to predict using input across multiple timestamps. This is not covered in previous CGL benchmarks \cite{ZhangCGLB} \cite{ko2022begin}. We benchmark different CGL methods on two datasets, covering primitive movements to daily activities.
    
    \item \changedr{We extend the order sensitivity issue proposed by \cite{yoon2019scalable} to two different setting: \textbf{task-order sensitivity} and \textbf{class-order sensitivity}. By comparing the result of both settings, we can capture the imbalance of performance of the classes within the same task as in \cref{fig:tsk_example}.}
    
    \item We present extensive empirical experiments on the order and architectural sensitivity of well-known CGL methods. I.e., the performance fluctuation \textbf{for each task/class} when the learning order changes, and the performance fluctuation when the backbone architecture changes. We are the first to study the \textbf{class-order sensitivity} of the CGL methods. We demonstrate that task-order robust methods can still be class-order sensitive. The scale of our experiment on order sensitivity is larger than in any previous works. Our work provides a comprehensive view of the order sensitivity of the experimented CGL methods and a setup for benchmarking CGL methods' sensitivity in class-IL on skeleton-based action recognition.
    
    \item \changedr{We study the correlation between the two most used evaluation metrics in CGL: Average Forgetting (\AF{}) and Average Accuracy (\AAC{}). We propose a theorem that defines the upper bound of \AF{} when \AAC{} is given. We visualize this in our results.}
    
    \item We compare the result of our benchmark with previous empirical studies in CL and demonstrate the difference in architectural sensitivity when graph-structured data are used instead of Euclidean data.
\end{itemize}

\section{\uppercase{Related Works}}\label{Related}

\textbf{CGL benchmarks.} \cite{ZhangCGLB} created a benchmark for CGL methods based on public datasets. It includes both node-level and graph-level prediction tasks in class-IL or task-IL CGL. \cite{ko2022begin} extends the benchmark to include edge-level prediction tasks, as well as tasks that use different CGL setting, such as domain-incremental learning or time-incremental learning. However, the data used in these benchmarks do not contain the temporal dimension. Our benchmark implements skeleton-based action recognition datasets that contain spatio-temporal graphs. It requires the model to reason over multiple timestamps of human skeleton-joint coordinates for accurate prediction. We note that the time-incremental learning in \cite{ko2022begin} refers to creating tasks based on the timestamp of the data, the data itself do not contain a temporal dimension. Moreover, not all benchmarks consider different task/class orders. Our benchmark experiments with different orders and we report the order sensitivity.

\textbf{Task-order sensitivity.} \cite{yoon2019scalable} proposed the problem of task-order sensitivity, where the CL performance for each task fluctuates when the order of tasks changes. They defined the \textit{Order-normalized Performance Disparity (OPD)} metric to evaluate it and reported that the tested CL methods have a high task-order sensitivity. \cite{bell2022effect} studied the significance of task orders for CL performance. They used synthetic data to find the best task order by calculating the distance between the tasks and proposed a way to estimate task distances using datasets such as MNIST using gradient curvature. They found that reordering the tasks to create the largest total task distances yields better results. We note that we do not always have control over task order, thus, the study of task-order sensitivity is necessary. \cite{lin2023theory} theorized the catastrophic forgetting effect in task-IL. One of the conclusions is that the best task order for minimizing forgetting tends to arrange dissimilar tasks in the early stage of CL. This corresponds with \cite{bell2022effect}. Further, other works studied the impact of different task orders in task-incremental CL setting too \cite{de2021continual} \cite{li2022provable}. Our work distinguishes from them by considering the more difficult class-IL CGL. 

Except for \cite{yoon2019scalable}, these works only consider \changedr{the sensitivity of the average accuracy and average forgetting, calculated across the tasks, but not the order sensitivity defined in \cite{yoon2019scalable}, where the performance of each task is evaluated separately}. Our work additionally benchmarks this problem extensively, and on the more fine-grained class level, revealing another problem that real-world applicable CL methods need to solve.

\textbf{Class-order sensitivity.} Class-order sensitivity is the problem where the CL performance per class varies when the classes within each task change. \cite{masana2020class} experimented with different class ordering. They constructed task sequences with increasing or decreasing difficulty in class-IL. They found that, for different CL methods, optimal performance is achieved using different class ordering. \cite{he2022rethinking} studied with a dataset where classes can be grouped as superclasses. They reported that creating tasks with classes from different superclasses yields better CL performance. These two works only consider a limited number of class orders. Our work generates 100 random class orders, group classes adjacent in the generated order into tasks, and perform class-IL CGL based on the modified tasks to approximate the class-order sensitivity of the CGL methods. Next, these works aim to find the best class order to increase the CL performance but omit the evaluation of performance changes for each class. Our work considers the performance change per class to assess the fairness of the CGL method in real-world applications.

\textbf{Impact of network architecture on CL.} \cite{goodfellow2013empirical} showed that adding a dropout layer to feed-forward networks improves CL performance. \cite{de2021continual} expanded the research by additionally reporting the experimental results for 4 models with different widths and depths with CL methods on VGG and AlexNet-based architectures. They observed that (too) deep models do not have high accuracy, while wider models often perform better. However, a wider model has a higher risk of over-fitting on the first task. \cite{mirzadeh2022architecture} studied the impact of different architectures and building blocks on CL performance. They used existing state-of-the-art (SOTA) computer vision architectures such as ResNet \cite{he2016deepresnet} and Vision Transformers \cite{dosovitskiy2020image-vit}. They observed that wide and shallow networks often outperform thin and deep networks in CL. Our work conducts architectural sensitivity experiments using GNNs, which react differently for transfer learning \cite{hu2019strategies}. \changedr{We also report the evolution of the performances with growing depth and width.}

\textbf{Continual Action Recognition.} To the best of our knowledge, Else-Net \cite{li2021elsenet} is the only previous literature which studied CGL with action recognition. They used the domain-incremental setting, where the subject / viewpoint is used to create tasks. This is a simpler setting as each task includes data samples of all classes. Our work evaluates CGL methods in the task and class incremental setting, where each task only contains unseen human actions.

\section{\uppercase{Preliminaries}}

\subsection{Class-incremental Learning}
We use a similar definition of class-IL like \cite{zhou2023deep}. Class-IL aims to learn from an evolutive stream with incoming new classes \cite{rebuffi2017icarl}. We assume a sequence of $B$ tasks: $\{D^1,D^2,D^3,...D^B\}$. $D^b = \{(x^b_i,y^b_i)\}_{i=1}^{n_b}$ denotes the $b$-th incremental training task with $n_b$ graph-structured data instances. Each task $D$ contains data instances from one or more classes $c \in C$. We denote function $Cls$ as a function that maps the task to the classes of the data instances it contains. The classes in each task $D$ do not overlap with any other tasks defined in the sequence. Thus:
\begin{equation}
Cls(D^a) \cap Cls(D^b) = \emptyset, \forall a,b \in \{1,...,B\}, a\neq b    
\end{equation}
During the $b$-th incremental training process, the model only has access to the data instances from the current task $D^b$. After each incremental training process, the model is evaluated over all seen classes $C_{seen}^b = Cls(D^1) \cup ... \cup Cls(D^b)$. Class-IL aims to find the model $f(x):X \mapsto C$ which minimizes the loss over all the tasks in the task sequence.

\subsection{Metrics} \label{Metrics}
We define \AAC{} and \AF{} as in \cite{chaudhry2018riemannian} and the \OPD{} metrics as in \cite{yoon2019scalable}, which is used to measure order sensitivity.

\textbf{Average Accuracy (\AAC{})}: The accuracy $a_{k,j} \in [0,1]$ is evaluated on the test set of task $j$ after training the model from task $1$ to $k$. The average accuracy of the model after incrementally training up to task $k$ is: 
\begin{equation}
\label{eqn:AA}
    \AAC{_k}=\frac{1}{k}\sum^{k}_{j=1}a_{k,j}
\end{equation}

\textbf{Average Forgetting (\AF{})}:
The average forgetting metric is defined on top of the average accuracy metric. First, we define forgetting for a single task $j$ as:
\begin{equation}
\label{eqn:F}
f_j^k=\max_{l\in\{1,...,k-1\}} a_{l,j}-a_{k,j}, \;\; \forall j<k 
\end{equation}
$f_j^k\in[-1,1]$ denotes the forgetting of task $j$ after learning task $k$. Next, we define the average forgetting for a model after learning task $k$ as: $\AF{_k}=\frac{1}{k-1}\sum^{k-1}_{j=1} f_j^k$.

\textbf{Order-normalized Performance Disparity (\OPD{})}:
order-normalized performance disparity is defined as the disparity between the model performance for a task $k$ on $R$ random task orders. I.e., it measures the impact of the task order on the performance of each task. Following \cite{yoon2019scalable}, we use accuracy to denote the model's performance:
\begin{equation}
\label{eqn:OPD}
\OPD{_t}=\max(a_{B,t}^{1},...,a_{B,t}^{R}) - \min(a_{B,t}^{1},...,a_{B,t}^{R})
\end{equation}
Where $a_{B,t}^{r}$ denotes the accuracy of task $t$ after learning the full task sequence of length $B$ in the random task order $r$. The maximum \OPD{} (\MOPD{}) is defined as $\MOPD{} = \max(OPD_1,...,OPD_B)$ and the average \OPD{} (\AOPD{}) is defined as $\AOPD{} =\frac{1}{B}\sum_{t=1}^{B}OPD_t$. We use this metric to report the task-order sensitivity. We adopt the same formula for class-order sensitivity, but instead of using task-wise accuracy, we will use the accuracy of the individual class. 

\subsection{Continual Learning Methods}
We used popular CL/CGL methods for our experiments. We refer to \cite{febrinanto2023graph} for the taxonomy.

Regularization-based methods prevent catastrophic forgetting by preserving weights from the previous model based on their importance. We have chosen \EWC{} \cite{kirkpatrick2017overcoming-ewc}, \MAS{} \cite{aljundi2018memory-mas} and \TWP{} \cite{liu2021overcoming-twp}. \EWC{} approximates the weight importance using the Fisher information matrix. \MAS{} uses the sensitivity of the output to a change in weight as importance. \TWP{} is designed specifically for CGL, it extends \MAS{} by additionally measuring the sensitivity of neighborhood topology changes when computing importance. Knowledge distillation-based methods preserve the output from the previous model. We have chosen \LWF{} \cite{li2017learning-lwf} for the experiments. 

Rehearsal-based methods preserve past data instances and replay them frequently during the learning of new tasks. We have chosen \GEM{} \cite{lopez2017gradient-gem}, which is a hybrid of rehearsal and regularization-based methods. \GEM{} stores a part of past data samples and regularizes the gradient to ensure that the loss of the stored samples does not increase.

Next, we implement a simple, purely rehearsal-based CGL method. It stores $x$\% of the data instance of each task as a memory buffer. The data instances in the buffer are mixed with the data instances from the new tasks and replayed to the model. This strategy is denoted as \REPLAY{} in the results of our experiments. We report the hyperparameter, the search-strategy, and the run times of the methods in Appendix \hyperref[C]{C}.

Architectural-based methods modify the architecture dynamically to prevent catastrophic forgetting. As one of our focuses is architectural sensitivity, we did not include architectural-based methods in our study.

Finally, we implement the \BARE{} and \JOINT{} baseline. \BARE{} baseline trains a deep learning model without any CGL method, models trained by this baseline will suffer catastrophic forgetting. It is used as the lower bound. \JOINT{} baseline trains a deep learning model by accumulating all available past data, it is equivalent to \REPLAY{} which stores 100\% of the data as buffer. \JOINT{} baseline is used as the upper bound. 

\section{\uppercase{Experiment Setup}} \label{Experiments}
In this section, we introduce our experimental setup, including the datasets, our experiments, and their goals. We provide the parameters for each experiment in Appendix \hyperref[B]{B} and Appendix \hyperref[C]{C}.

\subsection{Visualization}
\begin{figure}[ht]
  \centering
  \includegraphics[width=0.55\columnwidth]{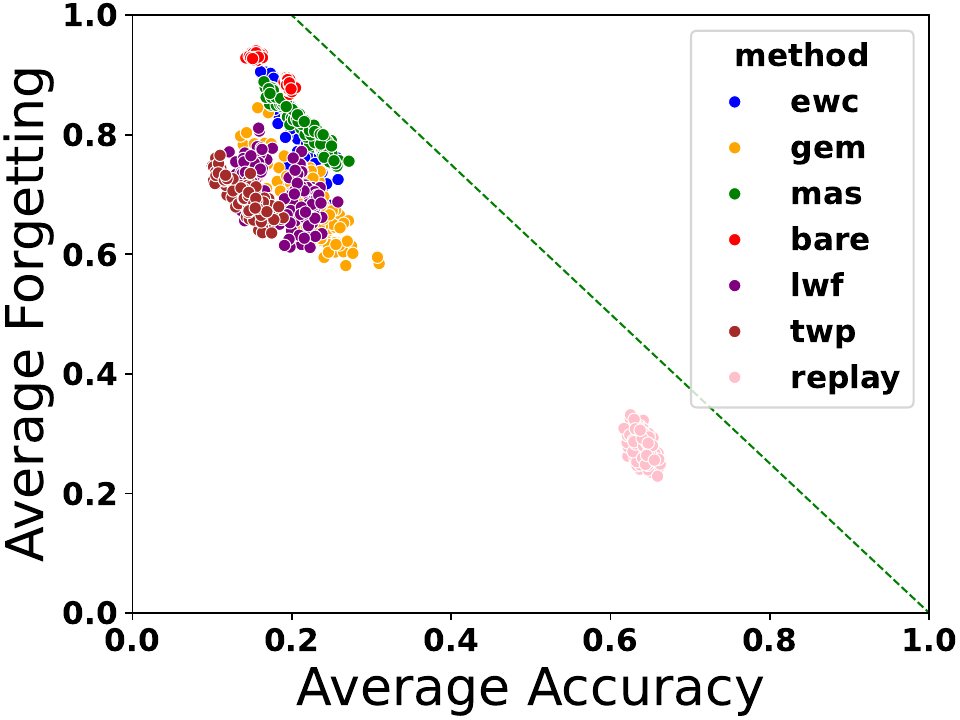}
  \caption{Example scatter plot of \AAC{} and \AF{}. The dotted green diagonal line shows the theoretical upper bound of \AF{}.
  \label{fig:metric_explain}}
\end{figure}

To facilitate compact sharing of our experiment results, we visualize them using the scatter plot with \AAC{} on the x-axis and \AF{} on the y-axis. This visualization is advantageous compared to standard mean and variance reporting, as we can observe the exact distribution across different experiments. E.g, in \mbox{\cref{fig:metric_explain}}, we observe two clusters of red points, one cluster has lower \AAC{} as they end with a difficult task. \changedr{Moreover, we visualize the theoretical upper bound of \AF{} for any given \AAC{} by proposing \cref{theorem:DD}. This shows that an improvement of \AAC{} will also improve \AF{} by lowering its upper bound. The proof is provided in Appendix \hyperref[A]{A}.}

\begin{theorem}
\label{theorem:DD}
Let \AAC{_k} be the average accuracy of the model after incrementally learning up to task \textbf{k} in class-IL. Then, the following inequation denotes the upper bound of \AF{_k}.
\begin{equation}
\label{eqn:5}
    \AF{_k} \le 1 - \frac{k}{k-1}\AAC{_k} + \frac{1}{k-1} a_{k,k}
    \end{equation}
\end{theorem}

\subsection{Datasets}
We implement two different action recognition datasets on our benchmark to cover different complexity of classes and tasks. N-UCLA \cite{wang2014cross-ucla} covers primitive movements, e.g. walk, pick up. NTU-RGB+D \cite{shahroudy2016ntu} covers daily activities, e.g. drink water, eat meal. We only used the first 10 classes from NTU-RGB+D for the construction of our class-IL CGL variant of the dataset so that both dataset has the same number of classes and tasks. We use the data processing protocol in \cite{chen2021channel-ctr} for both datasets.

We created tasks based on the classes of the data samples. Each task contains two classes and is divided into train, val, and test set with 8:1:1 ratio. The detailed information of the datasets is presented in \cref{tab:data}. The dataset will be released to facilitate future research.

\subsection{Backbone Graph Neural Networks}
Most of the CL/CGL methods require a backbone neural network. We use GCN \cite{kipf2016semi-gcn} \changedr{and ST-GCN \cite{yan2018spatial} as backbone GNNs. GCN is the foundational GNN that is used for many SOTA graph learning tasks. ST-GCN is a specialized GNN for the task of skeleton-based action recognition.} Appendix \hyperref[B]{B} contains the implementation details.

\begin{table}[ht]
\caption{Constructed class-incremental CGL dataset} \label{tab:data} 
\begin{center}
{\small{
\begin{adjustbox}{width=\columnwidth, center}
    \begin{tabular}{lcc}
    \toprule
    Dataset         & UCLA-CIL & NTU-CIL \\ 
    \midrule
    Data source                & \makecell{N-UCLA \\\cite{wang2014cross-ucla}}    & \makecell{NTU-RGB+D \\\cite{shahroudy2016ntu}}    \\ 
    \midrule
    \# action seq.                  & 1484          & 6667             \\ 
    \# joints        & 20         & 25            \\ 
    \# classes                 & 10         & 10            \\ 
    \# tasks                   & 5          & 5             \\ 
    \midrule
    avg. \# seq./task  & 297          & 1333             \\
    \# frame/seq. & 52          & 300             \\
    \bottomrule
    \end{tabular}
\end{adjustbox}
}}
\end{center}
\end{table}

\subsection{Experiments}
We introduce our experiments with their objectives and present the results in \cref{results}.

\subsubsection{Order Sensitivity}\label{order_sensitivity_explain}
The goal of the order sensitivity experiments is to test the \changedr{variance of CGL performance for each task individually when the order of tasks or classes is undetermined. The performance of each task is measured by \AAC{}, and the order sensitivity is measured by \OPD{}. }

We execute two types of experiments to test the order sensitivity empirically. In the task-order sensitivity experiment, we randomly shuffle the order in which the tasks are presented, without perturbing the class that constructs the tasks. In the class-order sensitivity experiment, we randomly shuffle the order in which the classes are presented to construct the tasks. I.e., tasks are constructed with random classes. This tests both the impact of task learning order and the impact of task difficulty. The class-order sensitivity experiment is more difficult and closer to real-world setting, \changedr{as real-world data often arrive in random order.} \changedr{The \OPD{} of task-order sensitivity is computed via task-level average accuracy, while the \OPD{} of class-order sensitivity is computed via the accuracy of each class.}

We note that a task order can be transformed into a class order. E.g., a task order $\{1,2,3,4,5\}$ is equivalent with the class order $\{0,1,2...,8,9\}$. Thus, we aggregated the results of both experiments to obtain a more accurate approximation of class-order sensitivity.

\subsubsection{Architectural Sensitivity}
The architectural sensitivity experiment aims to test the performance stability of the CGL methods with different widths and depths of the backbone network. We measure this by observing the \changedr{evolution} of \AAC{} and \AF{} \changedr{with different GNN architecture}.

Although previous studies \cite{de2021continual} \cite{mirzadeh2022architecture} report that a wide and shallow architecture often outperforms thin and deep architectures, the evolution of performance changes is never reported. \changedr{We experiment with a gradual change in model width and depth} and report the results for each step. This information is useful to determine \changedr{the trend} of performance changes.

\section{\uppercase{Results}}\label{results}

\subsection{Task-order Sensitivity}
\begin{figure}[ht]
  \centering
  \begin{subfigure}[b]{0.49\columnwidth}
      \centering
      \includegraphics[width=\columnwidth]{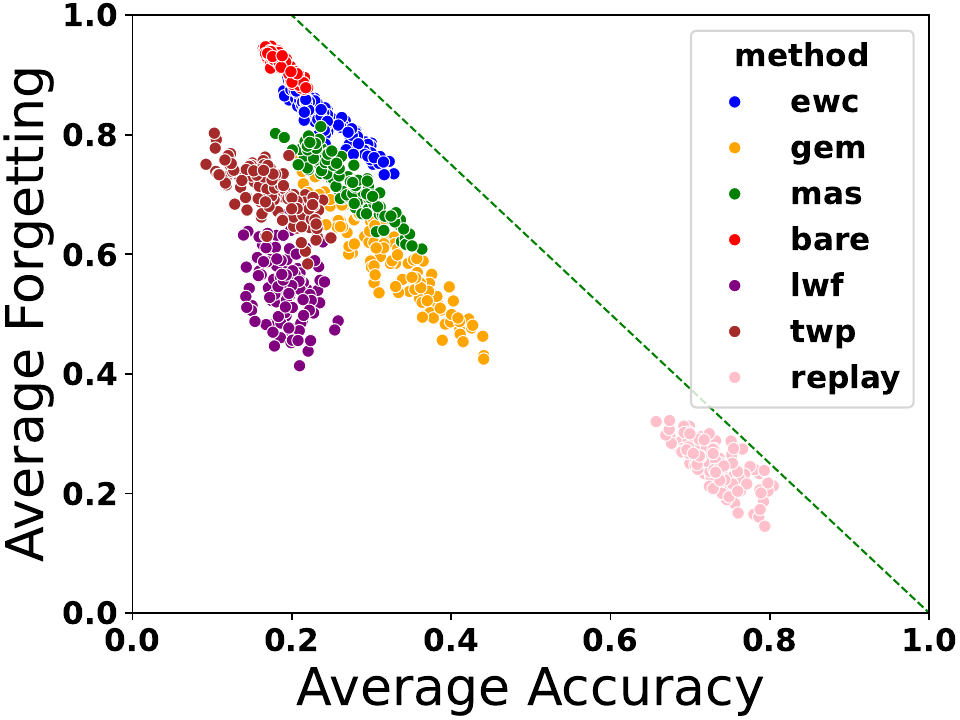}
      \caption{UCLA-CIL} \label{fig:tsk_scatter_ucla}
  \end{subfigure}
  \hfill
  \begin{subfigure}[b]{0.49\columnwidth}
      \centering
      \includegraphics[width=\columnwidth]{support_figures/tsk_NTU_GCN_scatter.pdf}
      \caption{NTU-CIL}
  \end{subfigure}
  
  \caption{Scatter plot of \AAC{} and \AF{} for task-order experiment \changedr{ with GCN}.}
  \label{fig:tsk_scatter}
\end{figure}

\begin{figure}[ht]
  \centering
  \begin{subfigure}[b]{0.49\columnwidth}
      \centering
      \includegraphics[width=\columnwidth]{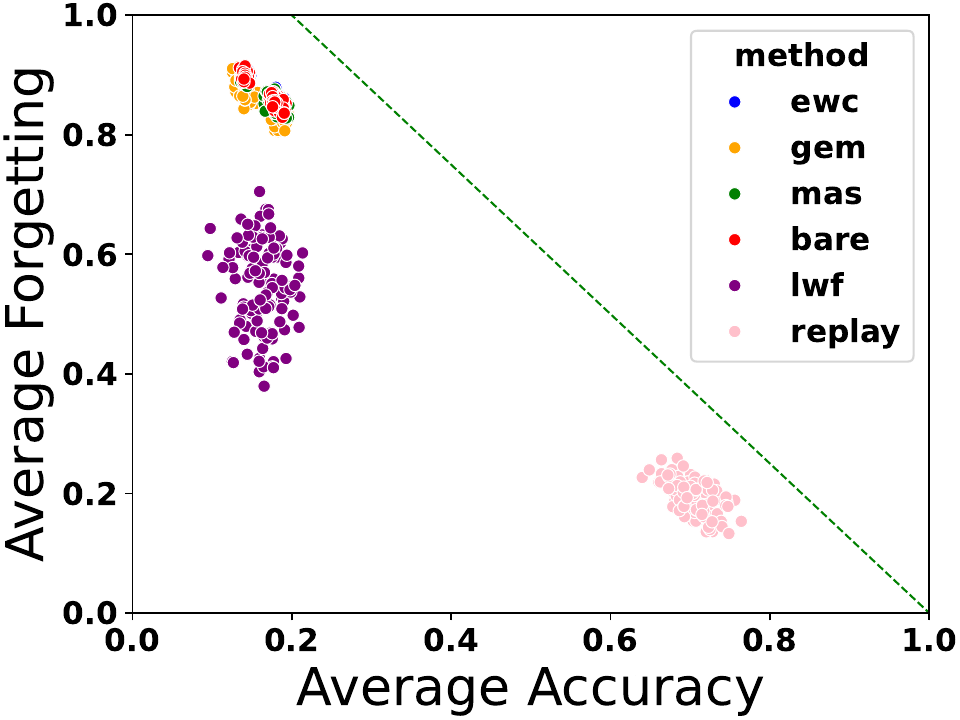}
      \caption{UCLA-CIL}\label{fig:tsk_scatter_STGCN_UCLA}
  \end{subfigure}
  \hfill
  \begin{subfigure}[b]{0.49\columnwidth}
      \centering
      \includegraphics[width=\columnwidth]{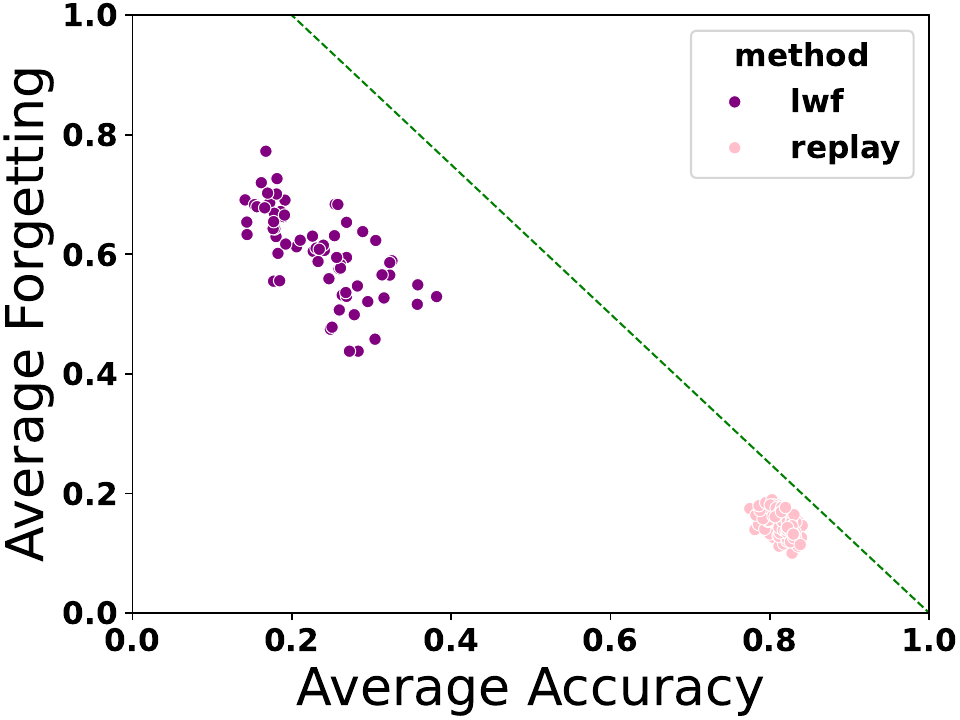}
      \caption{NTU-CIL}
  \end{subfigure}
  
  \caption{Scatter plot of \AAC{} and \AF{} for task-order experiment \changedr{ with ST-GCN}.}
  \label{fig:tsk_scatter_STGCN}
\end{figure}

We first discuss the task-order sensitivity experiment. As we only have $5$ tasks for our datasets, we have $5!=120$ different task orders. We experimented with all $120$ task orders \changedr{with GCN as backbone} and visualized their \AAC{} and \AF{} in \cref{fig:tsk_scatter}. \changedrg{The result of ST-GCN are visualized in  \cref{fig:tsk_scatter_STGCN}. We restrain ourselves of extensively experimenting ST-GCN on the NTU-CIL dataset, as all the regularization-based methods performed on-par with the \BARE{} baseline in the simpler UCLA-CIL dataset. Our results are sufficient to support the hypothesis. The explanation given below applies to both backbones as the results share similar trends}. We note that order sensitivity does not apply to the \JOINT{} baseline, as \JOINT{} will be training on all available data in the end, changing the class/task order will have little effect. Thus, we did not train \JOINT{} for order sensitivity experiments.

From \cref{fig:tsk_scatter}, we observe clear clusters for different methods in UCLA-CIL. Moreover, \changedr{different methods exhibit different characteristics}. Regularization-based methods are sensitive for the performance metric \AAC{}, while \LWF{} and \REPLAY{} are more stable. Interestingly, we see that \GEM{} performed significantly worse and is more sensitive compared to \REPLAY{}, even though they store the same fraction of the past data as buffer. This could mean that regularization-based methods do not prevent catastrophic forgetting as well as rehearsal-based methods in skeleton-based action recognition, where the data from the different action classes looks vastly different. We hypothesize that none of the CGL methods learns features in the first task that can be generalized for the full task space. However, \REPLAY{} can escape the local minimum faster because it does not rely on the output or weights of previous models. This leads to better generalization and a higher \AAC{} of \REPLAY{}. \changedr{This phenomenon can also be observed in \cref{fig:tsk_scatter_STGCN_UCLA}. All the regularization based methods have low \AAC{} as the embeddings they learn are not generalizable to other tasks.}

\begin{table*}
\begin{center}
    {\small{
    
\caption{Order sensitivity measured by OPD \changedr{ for experiments with a GCN backbone}. Bold entries are the lowest OPDs, underlined entries are the highest OPDs, excluding the \BARE{} baseline. The OPD metric in the class-order sensitivity experiment is based on the accuracy of the classes. \changedr{All the methods are order sensitive, there is a large difference in task-order sensitivity and class-order sensitivity.}}
\label{tab:mopds}
\begin{adjustbox}{width=\textwidth, center}
    \begin{tabular}{ccccc|cccc}
    \toprule
     & \multicolumn{4}{c|}{task-order sensitivity experiment} & \multicolumn{4}{c}{aggregated class-order sensitivity experiment}\\
    method\textbackslash Setting& \multicolumn{2}{c}{UCLA-CIL} &\multicolumn{2}{c|}{NTU-CIL} & \multicolumn{2}{c}{UCLA-CIL} &\multicolumn{2}{c}{NTU-CIL}\\
    
     &\AOPD($\downarrow$) & \MOPD($\downarrow$)&\AOPD($\downarrow$) & \MOPD($\downarrow$)&\AOPD($\downarrow$) & \MOPD($\downarrow$)&\AOPD($\downarrow$) & \MOPD($\downarrow$)\\
    \midrule
       \BARE{}    
       & 96.74\%$\pm$0.72\%
       & 100.0\%$\pm$0.0\%
       & 92.91\%$\pm$0.38\%
       & 99.24\%$\pm$0.0\%
       & 100.0\%$\pm$0.0\%
       & 100.0\%$\pm$0.0\%
       & 99.73\%$\pm$0.06\%
       & 100.0\%$\pm$0.0\%\\
   \midrule
       \EWC{}     
       & \underline{93.94\%$\pm$0.56\%}
       & \underline{100.0\%$\pm$0.0\%}
       & 85.36\%$\pm$0.88\%
       & \underline{94.24\%$\pm$1.23\%}
       & \underline{100.0\%$\pm$0.0\%}
       & \underline{100.0\%$\pm$0.0\%}
       & \underline{98.12\%$\pm$0.49\%}
       & \underline{100.0\%$\pm$0.0\%}\\
       \MAS{}     
       & 89.46\%$\pm$1.6\%
       & 98.7\%$\pm$1.61\%
       & \underline{87.88\%$\pm$0.51\%}
       & 95.61\%$\pm$1.3\%
       & 99.73\%$\pm$0.33\%
       & \underline{100.0\%$\pm$0.0\%}
       & 97.61\%$\pm$0.4\%
       & \underline{100.0\%$\pm$0.0\%}\\
       \TWP{}     
       & 85.39\%$\pm$1.93\%
       & \underline{100.0\%$\pm$0.0\%}
       & 77.45\%$\pm$1.4\%
       & 89.24\%$\pm$1.11\%
       & 99.3\%$\pm$0.45\%
       & \underline{100.0\%$\pm$0.0\%}
       & 95.73\%$\pm$0.74\%
       & \underline{100.0\%$\pm$0.0\%}\\
       \LWF{}     
       & 63.21\%$\pm$2.46\%
       & \textbf{75.04\%$\pm$6.46\%}
       & 78.36\%$\pm$1.1\%
       & 93.48\%$\pm$1.13\%
       & 93.44\%$\pm$1.46\%
       & \underline{100.0\%$\pm$0.0\%}
       & 94.0\%$\pm$0.61\%
       & 98.79\%$\pm$0.61\%\\
       \GEM{}     
       & 83.89\%$\pm$1.45\%
       & \underline{100.0\%$\pm$0.0\%}
       & 80.91\%$\pm$0.25\%
       & 94.09\%$\pm$0.88\%
       & 99.59\%$\pm$0.33\%
       & \underline{100.0\%$\pm$0.0\%}
       & 95.36\%$\pm$0.49\%
       & \underline{100.0\%$\pm$0.0\%}\\
       \REPLAY{}  
       & \textbf{58.21\%$\pm$7.54\%}
       & 80.25\%$\pm$5.2\%
       & \textbf{39.52\%$\pm$0.94\%}
       & \textbf{53.18\%$\pm$2.41\%}
       & \textbf{82.54\%$\pm$4.56\%}
       & \underline{100.0\%$\pm$0.0\%}
       & \textbf{50.52\%$\pm$1.16\%}
       & \textbf{71.82\%$\pm$2.27\%}\\
    \bottomrule
    \end{tabular}
\end{adjustbox}
    }}
\end{center}
\end{table*}

Next, \changedr{as proven by \cref{theorem:DD}, changes in \AAC{} lowers the upper bound of \AF{}. From \cref{fig:tsk_scatter_ucla}, we observe a linear correlation between the increase in \AAC{} and the decrease in \AF{}, this hints the correlation between \AAC{} and \AF{}.} However, we observe in \cref{fig:tsk_scatter_ucla} different levels of \AF{} \changedr{when the \AAC{} of the different methods are similar. We suspect that it indicates regularization strength of the different regularization-based CGL methods under the scenario when the embeddings are not generalizable, i.e., when the model overfits to previous tasks.} \MAS{} computes the sensitivity of the output to each weight. When the model overfits, the output will likely be very sensitive to a subset of weights. \MAS{} regularizes the model more on these weights, hindering the learning of new tasks. \changedr{This reduces the maximum \AAC{} of the new tasks, thus reducing \AF{}.} In NTU-CIL, where overfitting is less likely due to a larger sample size, \AF{} of \MAS{} is comparable to \EWC{}. \TWP{} has an extra regularization on top of \MAS{} that preserves the topology. However, SOTA GNNs on skeleton-based action recognition \cite{chen2021channel-ctr} show that using different topologies is beneficial to prediction accuracy. Thus, the \AF{} of \TWP{} is lower than \MAS{} when their \AAC{} are similar. Finally, \GEM{} prevents the loss increment of the stored samples. When model overfits to past tasks, \GEM{} is unable to unlearn it. This is reflected in the lowest \AF{} in both datasets. In the end, \LWF{}'s \AF{} is very sensitive to the task order. The features learned in the first task influence the learning of following tasks heavily. When the difference between the tasks is large, the features learned by the old model may not be important for the new task. In that case, the output of the old model can be unstable. Thus, the regularization strength of \LWF{} depends on the similarity of past and current tasks. A change in task order will influence this, and indirectly influence \AF{} too. \changedrg{The highest \AF{} for \LWF{} is obtained by the task order that starts with tasks with many leg movements, and ends with tasks that have minimal leg movements, while the task order with the lowest \AF{} has to alternate between tasks with many or minimal leg movements.} \changedr{We note that, a low \AF{} can be achieved by lowering the maximum \AAC{} or raising the minimum \AAC{} of each task during the CGL process. The phenomenon here corresponds to the former case. It does not improve the usability of the CGL model and is undesirable.}

Then, we study the task-order sensitivity by observing the \OPD{} metric. \OPD{} is calculated over 120 random task-orders and 220 random class-orders, we repeated the experiment 5 times, obtaining 5 \OPD{} metrics. We report the mean and standard deviation in \cref{tab:mopds}. Regularization-based methods are highly task-order sensitive in UCLA-CIL, while \LWF{} and \REPLAY{} are comparatively less sensitive. \EWC{}, \MAS{} and \GEM{} achieved 100\% \MOPD{}, which denotes that there exist two task orders and a task $A$, where task $A$ has 100\% \AAC{} in one order, and 0\% in another order. This behavior is not desirable for CL methods. As regularization-based methods only constrain the learnable weights, it is expected that the last learned tasks will have higher performances. When the tasks are sufficiently different, the weight will shift far enough so that the older tasks are forgotten. In comparison, \LWF{} and \REPLAY{} alter the gradient of the model by resp. knowledge distillation and rehearsal. By modifying the gradient using other means than constraints, the model has more freedom and may find a loss region where it generalizes to both present and past tasks.

In NTU-CIL, we see an improvement in task-order robustness for many methods, except \LWF{}. Increased task complexity decreases the chance of overfitting, which facilitates the regularization process and \REPLAY{} as they learn generalizable embedding that can be reused by following tasks. We observe two distinguishable clusters for \LWF{} and \BARE{} baseline, when we compare the task orders which form the two clusters, we find them to be identical. Moreover, the cluster with higher \AAC{} always ends with task 0 or 1. The classes for these tasks are resp. \{drink water, eat meal\} and \{brush teeth, brush hair\}. \changedr{Based on our earlier hypothesis, large difference between tasks hinders \LWF{} the most. In NTU-CIL, both primitive movement and daily actions exists, thus} the models that learned the primitive movement may not focus on the finer movements that identifies daily actions. This makes the knowledge distillation between old and new models unstable and \changedr{potentially destructive during the learning process of the new task}, which makes \LWF{} order sensitive.

\subsection{Class-order Sensitivity}
\begin{figure}[ht]
  \centering
  \begin{subfigure}[b]{0.49\columnwidth}
      \centering
      \includegraphics[width=\linewidth]{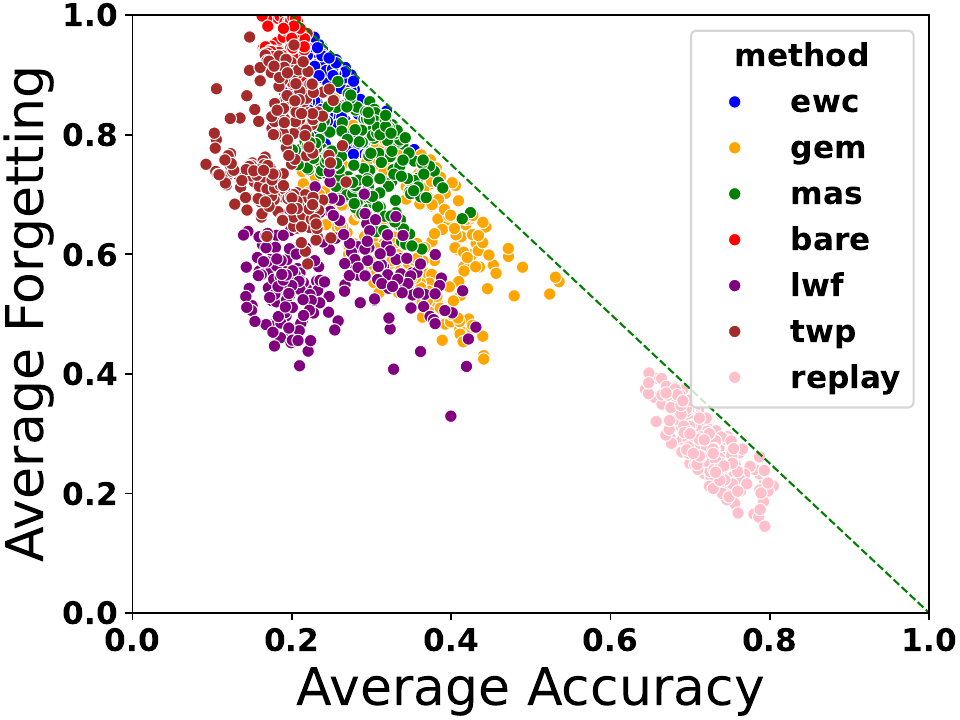}
      \caption{UCLA-CIL}
      \label{fig:cls_nucla_scatter}
  \end{subfigure}
  \hfill
  \begin{subfigure}[b]{0.49\columnwidth}
      \centering
      \includegraphics[width=\columnwidth]{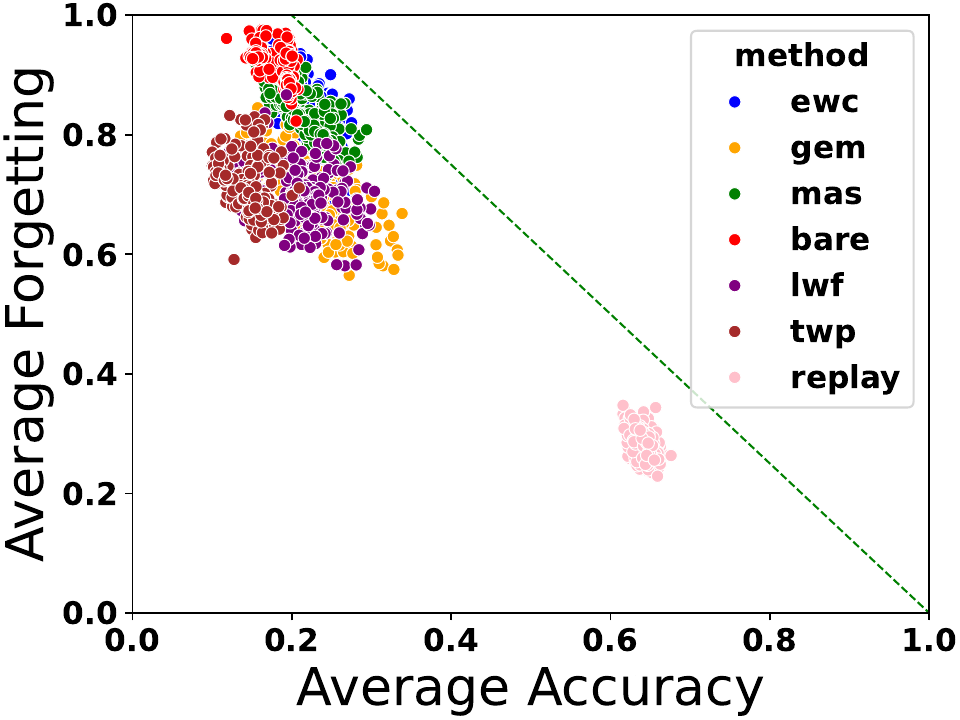}
      \caption{NTU-CIL}
  \end{subfigure}
  \caption{Scatter plot of \AAC{} and \AF{} for class and task-order experiments \changedr{with GCN}.}
  \label{fig:cls_scatter}
\end{figure}

\begin{figure}[ht]
  \centering
  \begin{subfigure}[b]{0.49\columnwidth}
      \centering
      \includegraphics[width=\linewidth]{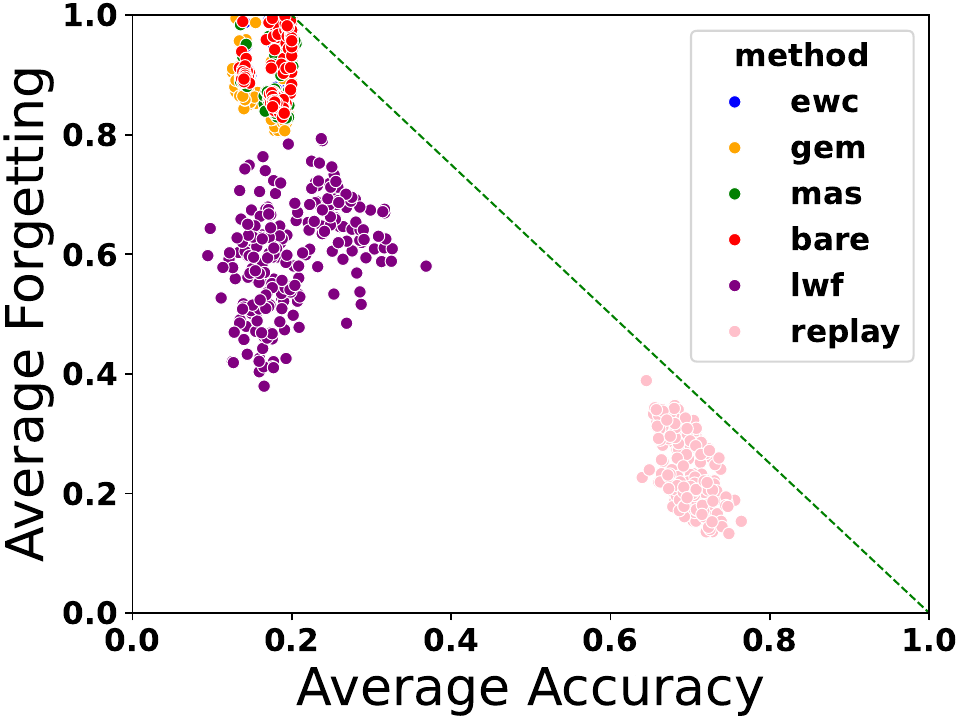}
      \caption{UCLA-CIL}
  \end{subfigure}
  \hfill
  \begin{subfigure}[b]{0.49\columnwidth}
      \centering
      \includegraphics[width=\columnwidth]{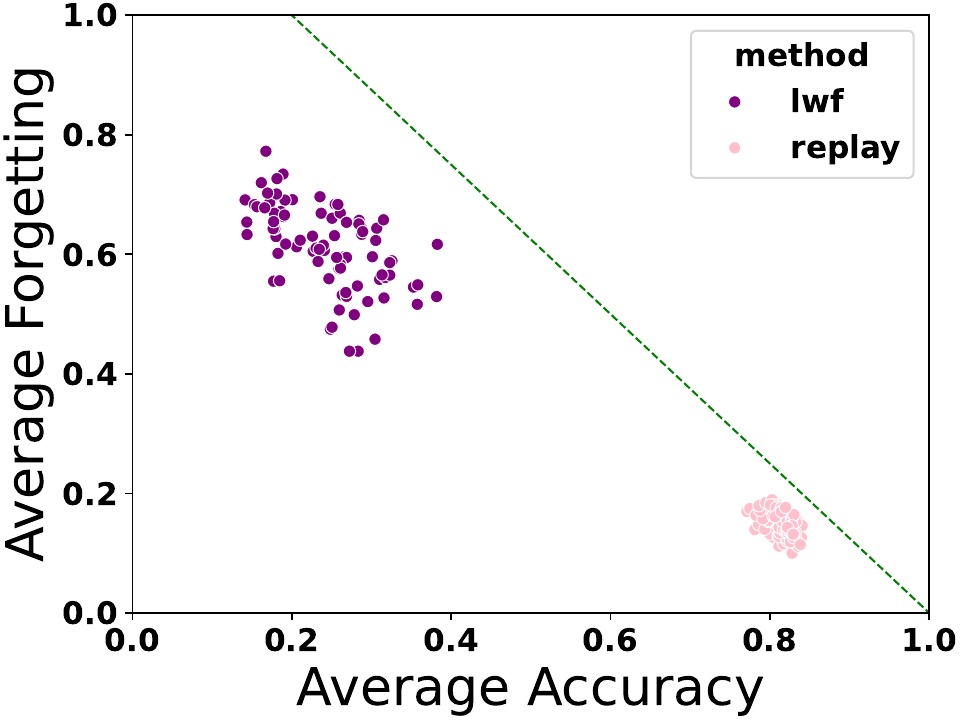}
      \caption{NTU-CIL}\label{fig:cls_ntu_scatter_stgcn}
  \end{subfigure}
  \caption{Scatter plot of \AAC{} and \AF{} for class and task-order experiments \changedr{with ST-GCN}.}
  \label{fig:cls_scatter_STGCN}
\end{figure}

We test 100 different class orders for each CGL method in both datasets. We aggregate the results from both experiments to obtain a more accurate class-order sensitivity of the CGL methods as mentioned in \cref{order_sensitivity_explain}, the metrics are visualized in \cref{fig:cls_scatter} and \cref{fig:cls_scatter_STGCN}. \changedr{\cref{fig:cls_ntu_scatter_stgcn} contains partial results, as mentioned above.}

\textbf{Limitation.} 100 random class order is only a small subset of all possible class orders, for our datasets, there will be $10!$ different class orders. It is not feasible to experiment on all the class orders, thus, our result only approximates the class-order sensitivity. 

\changedr{In \cref{fig:cls_scatter_STGCN}, we observe the same phenomenon where the regularization-based methods have a low \AAC{} for ST-GCN, which leads to the same hypothesis that non-generalizable embeddings are learned. In \cref{fig:cls_scatter}, we observe that, with randomized class order, the cluster of results is vastly different compared to the randomized task order. However, the trends are consistent with the previous experiment}: \EWC{} and \TWP{} perform the worst, \GEM{}, \MAS{} and \LWF{} perform better, and \REPLAY{} performs the best. Surprisingly, we see that all methods are class-order sensitive in UCLA-CIL by observing the \OPD{} metric in \cref{tab:mopds}. 

\changedr{The \OPD{} of task-order sensitivity is computed via task-level accuracy, while the \OPD{} of the class-order sensitivity is calculated using the accuracy of each class. A large difference between the two denotes that the accuracy of classes within the task are not equal. An example can be observed in \cref{fig:teaser}, in task $Y$, the accuracy of class 2 is much worse than the accuracy of class 3}. \changedr{Previous evaluation setting proposed by \cite{yoon2019scalable} can not detect this issue as the performance is evaluated at task level. Our evaluation setting is at the more fine-grained class level.} 

We hypothesize that this is caused by the unbalanced features used to identify the classes. The model may learn many features to identify one class and only a few features for the other classes. The latter class is more prone to catastrophic forgetting. Our results demonstrate that a task-order robust method can still have classes with bad performances, which is an issue in real-life settings where fairness is crucial. 

\subsection{Architectural Sensitivity}
\begin{figure}[ht]
  \centering
  \begin{subfigure}[b]{0.49\columnwidth}
  \centering
  \includegraphics[width=\linewidth]{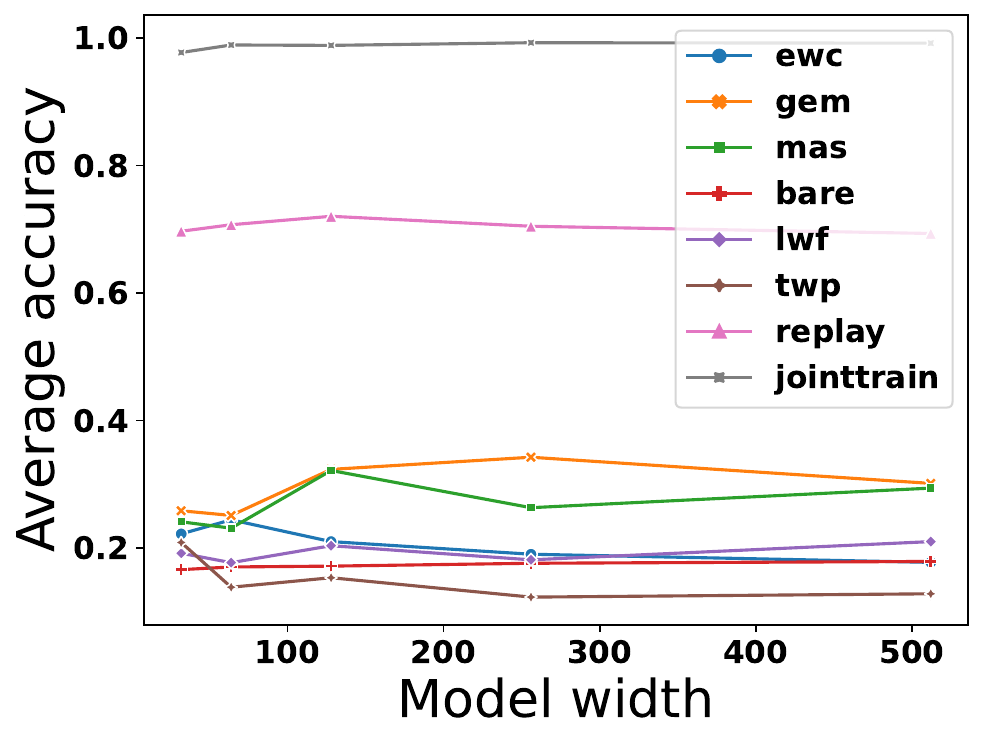}
  \caption{UCLA-CIL, \AAC{}}
  \end{subfigure}
  \hfill
  \begin{subfigure}[b]{0.49\columnwidth}
  \centering
  \includegraphics[width=\linewidth]{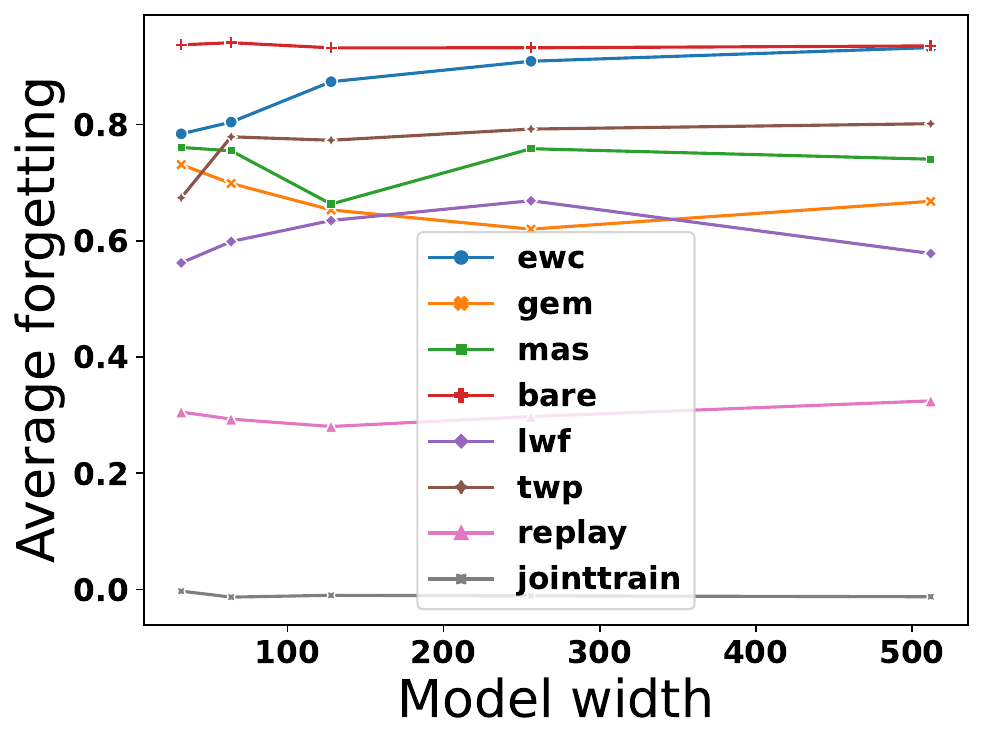}
  \caption{UCLA-CIL, \AF{}}
  \end{subfigure}
      \caption{Evolution of the metrics when GCN width varies.}
  \label{fig:arch_width}
\end{figure}

\begin{figure}[ht]
  \centering
  \begin{subfigure}[b]{0.49\columnwidth}
  \centering
  \includegraphics[width=\linewidth]{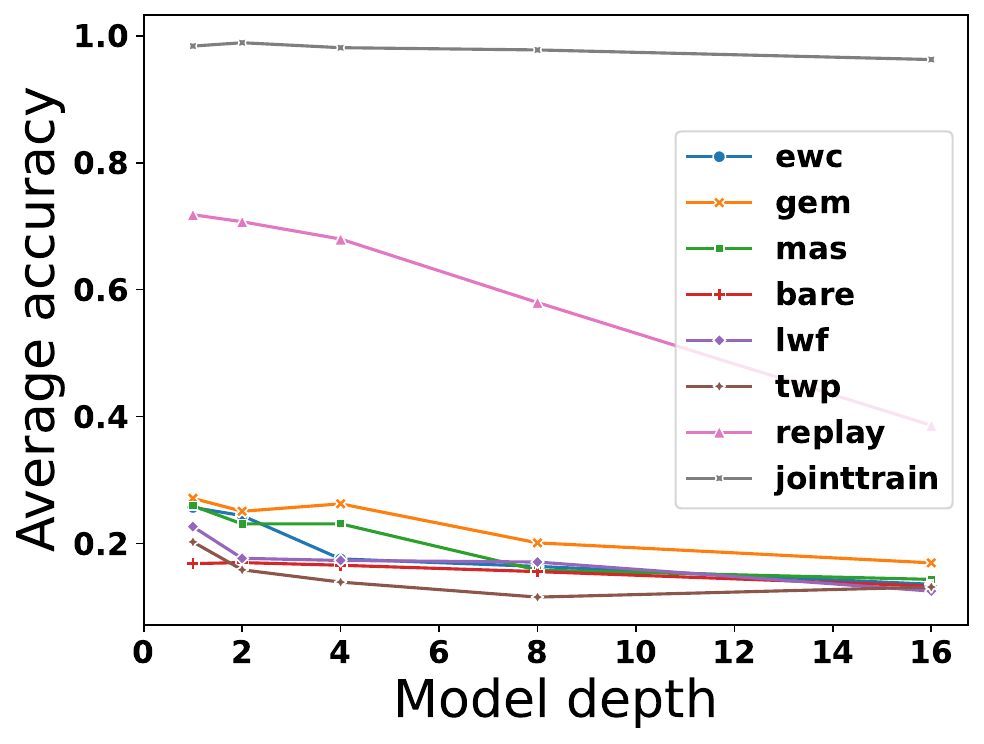}
  \caption{UCLA-CIL, \AAC{}}
  \end{subfigure}
  \hfill
  \begin{subfigure}[b]{0.49\columnwidth}
  \centering
  \includegraphics[width=\linewidth]{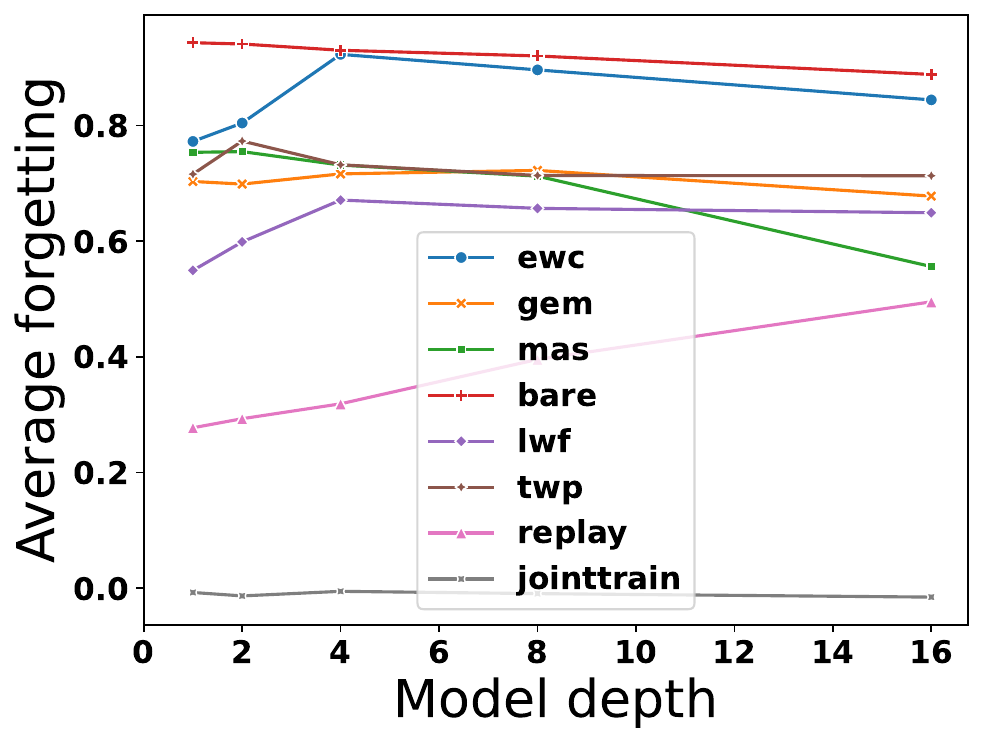}
  \caption{UCLA-CIL, \AF{}}\label{fig:depth_af}
  \end{subfigure}
  \caption{Evolution of the metrics when GCN depth varies.}
  \label{fig:arch_depth}
\end{figure}

We test the architectural sensitivity for the performance of the CGL models by increasing model width/depth \changedr{using a GCN as backbone}. We report the width evolution in \cref{fig:arch_width} and depth evolution in \cref{fig:arch_depth}.

Previous studies in CL \cite{de2021continual} \cite{mirzadeh2022architecture} establish that a wide and shallow network outperforms a thin and deep network. Our results with GNNs contradict their observations. Increasing the width does not always show gains, the effect of increasing depth is only consistent for \REPLAY{}.

We hypothesize that it is caused by the combination of GNN and skeleton-based action recognition tasks. The skeleton-joint graph input is connected sparsely. There are 10 edges between the left hand and left foot node. This makes shallow GNNs unable to capture high-level features that require information from both hand and foot movements, e.g., touching foot. Previous work \cite{dwivedi2022long} shows that GCN does not capture long-ranged information well. Our width sensitivity experiment uses 2-layer GNNs, which means that it can not produce the abovementioned features. Increasing the width only worsens the over-fitting of the tasks. This hypothesis also explains why increasing the depth does not always lead to worse performance, deeper GNNs can capture long-ranged information, creating more generalizable features, which increases the performance. \changedr{This can be observed in \cref{fig:depth_af}, we see a downward trend of \AF{} for most regularization-based methods as well as the \BARE{} baseline.} However, increasing the depth also worsens the overfitting effect due to the increased expressiveness. The over-squashing phenomenon also prevents the model from learning high-level features \cite{alon2020bottleneck}. This is most noticeable in \REPLAY{}, where the overfitting can occur on both the current task and on the buffers of the past tasks. Thus, contrary to observation in previous work, for skeleton-based action recognition, increasing the width is not useful for shallow GNNs, and increasing the depth may have some benefits. \changedr{This observation motivates future studies in deep-GNNs and their applications in the spatio-temporal graphs.}

\section{\uppercase{Conclusion}} \label{Discussion}
\changedr{We constructed the first Continual Graph Learning (CGL) benchmark for spatio-temporal graphs. We are the first to benchmark skeleton-based action recognition in class-incremental learning (class-IL) setting.}

\changedr{We extended the order sentivity issue proposed by \cite{yoon2019scalable} to two different settings: task-order sensitivity and class-order sensitivity. By comparing the order sensitivity at both task and class level, we captured the imbalance of performance between the classes within the same task, which is an unexplored problem in CGL.} We conducted extensive experiments on the task and class-order sensitivity of the popular CGL methods. We discovered that the task-order robust methods can still be class-order sensitive, \changedr{i.e., in some tasks, there are classes which outperforms the other classes}. \changedr{Next, we show that} popular CGL methods are all order sensitive, i.e. the performance of each task/class dependents on the learning order.

We also studied the architectural sensitivity. We report the evolution of \AAC{} and \AF{} when the depth and width of the backbone GNN varies. Our results contradict previous empirical observations in CL. We provided our insight on the contradiction.

\changedr{We studied the correlations between average forgetting (\AF{}) and average accuracy (\AAC{}), identified the upper-bound of \AF{}, demonstrated that an improvement of \AAC{} naturally lowers the upper-bound of \AF{} and visualized it in our results.}

The studies in order and architectural sensitivity are still underexplored in class-IL CGL. Our paper is an introduction to these two issues. Future works are 1) Expand the benchmark to include node-level and edge-level tasks. 2) Investigate the intuitions to propose class-order robust CGL methods. We hope that our paper initiates further research on the sensitivities of CGL methods.

\section*{\uppercase{Acknowledgements}}
This research received funding from the Flemish Government under the “Onderzoeksprogramma Artificiële Intelligentie (AI) Vlaanderen” programme.

\bibliographystyle{apalike}
{\small
\bibliography{example}}

\section*{\uppercase{Appendix A: theorem 1}}\label{A}
To prove that the upper bound of \AF{} is correlated to \AAC{}, we use the definition from \cref{Metrics}:

\begin{equation}
\label{eqn:AA2}
    \AAC{_k}=\frac{1}{k}\sum^{k}_{j=1}a_{k,j}
\end{equation}

\begin{equation}
\label{eqn:F2}
f_j^k=\max_{l\in\{1,...,k-1\}} a_{l,j}-a_{k,j}, \;\; \forall j<k 
\end{equation}

\begin{equation}
\label{eqn:AF}
\AF{_k}=\frac{1}{k-1}\sum^{k-1}_{j=1} f_j^k    
\end{equation}

Next, we expand \cref{eqn:AF} with \cref{eqn:F}:

\[\AF{_k}=\frac{1}{k-1}(\sum^{k-1}_{j=1} \max_{l\in\{1,...,k-1\}} a_{l,j}-\sum^{k-1}_{j=1} a_{k,j})\]

When $a_{k,j}$ increases, the upper bound of \AF{_k} will be smaller. When we assume the forgetting is maximum, without modifying \AAC{_k}, i.e. $\max_{l\in\{1,...,k-1\}} a_{l,j}$ is always 1, we will have:
\begin{equation}
\label{eqn:6}
\AF{_k} \le (\frac{1}{k-1}\sum^{k-1}_{j=1} 1) -\frac{1}{k-1}\sum^{k-1}_{j=1} a_{k,j}
\end{equation}
\[\AF{_k} \le 1 -\frac{1}{k-1}\sum^{k-1}_{j=1} a_{k,j}\]

The term $\frac{1}{k-1}\sum^{k-1}_{j=1} a_{k,j}$ is very closely related to \AAC{}, which is $\frac{1}{k}\sum^{k}_{j=1} a_{k,j}$. We can denote the term $\frac{1}{k-1}\sum^{k-1}_{j=1} a_{k,j}$ as $x$ and transform the inequality:

\[\AAC{_k} = \frac{1}{k}\sum^{k}_{j=1} a_{k,j}\]
\[\AAC{_k} = \frac{k-1}{k}\frac{1}{k-1}(\sum^{k-1}_{j=1} a_{k,j} + a_{k,k})\]
\[\AAC{_k} = \frac{k-1}{k}\frac{1}{k-1}(\sum^{k-1}_{j=1} a_{k,j}) + \frac{k-1}{k}\frac{1}{k-1}a_{k,k}\]
\[\AAC{_k} = \frac{k-1}{k}x + \frac{1}{k}a_{k,k}\]
\[\frac{k}{k-1}(\AAC{_k} - \frac{1}{k}a_{k,k})= x \]
When we replace the term $x$ in the original inequality by the transformed \AAC{_k}, we get the following:
\[\AF{_k} \le 1 - (\frac{k}{k-1}(\AAC{_k} - \frac{1}{k}a_{k,k}))\]
\[\AF{_k} \le 1 - \frac{k}{k-1}\AAC{_k} + \frac{1}{k-1}a_{k,k}\]
This denotes the upper bound of \AF{_k} for a given \AAC{_k}. However, to visualize the upper bound in the scatter plot, it is not possible to include $a_{k,k}$ of each experiment. For simplicity, we can assume $a_{k,k}$ is always at maximum, i.e. 100\% accuracy. We can then draw the dotted green line as visualized in \cref{fig:metric_explain} using \cref{eqn:7}:
\begin{equation}
    \label{eqn:7}
    \AF{_k} \le 1 - \frac{k}{k-1}\AAC{_k} + \frac{1}{k-1}
\end{equation}

\section*{\uppercase{Appendix B: Implementation}}\label{B}

Our benchmark adapted the code from CGLB \cite{ZhangCGLB}. We extended the code to handle extra experiment options, including reordering classes, tasks, and options for changing backbone architecture and the computation of corresponding metrics. 

\begin{table*}[ht]
\begin{center}
    {\small{
\caption{Best hyperparameter candidates found via grid search for GCN. frac\_memories denotes the percentage of each task that GEM/REPLAY stores as buffer.} \label{tab:best_hyper}
\begin{tabular}{lcc}
\toprule
algorithm & \multicolumn{2}{c}{best hyper-parameter}  \\
& N-UCLA-CIL & NTU-CIL\\
\midrule
EWC    & memory\_strength: 1000000 & memory\_strength: 1000000\\ 
MAS    & memory\_strength: 100 & memory\_strength: 100 \\ 
TWP    & \makecell{lambda\_l: 10000; lambda\_t: 10000; beta: 0.01} & \makecell{lambda\_l: 100; lambda\_t: 10000; beta: 0.01}\\ 
LwF    & lambda\_dist: 1; T: 2 & lambda\_dist: 0.1; T: 2 \\ 
GEM    & \makecell{memory\_strength: 5; frac\_memories: 0.2}  & \makecell{memory\_strength: 5; frac\_memories: 0.2}\\ 
REPLAY & frac\_memories: 0.2 & frac\_memories: 0.2\\ 
\bottomrule
\end{tabular}
    }}
\end{center}
\end{table*}

\begin{figure}[ht]
  \centering
  \begin{subfigure}[b]{0.48\columnwidth}
  \centering
  \includegraphics[width=0.75\columnwidth]{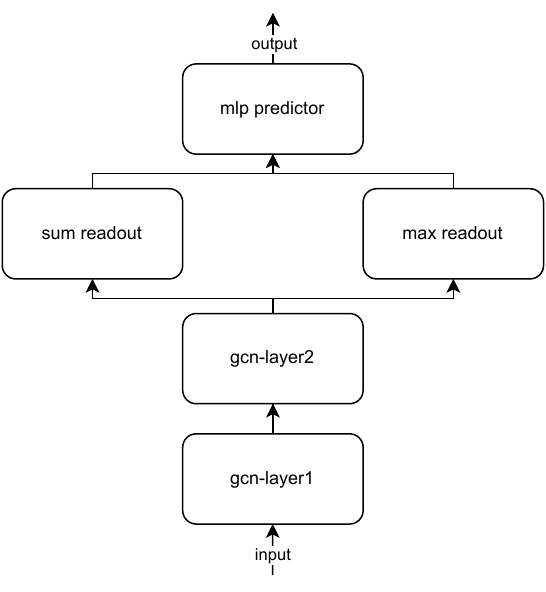}
  \caption{Visualization of baseline backbone GCN architecture.}
  \label{fig:gcn}    
  \end{subfigure}
  \hfill
  \begin{subfigure}[b]{0.48\columnwidth}
  \centering
  \includegraphics[width=0.76\columnwidth]{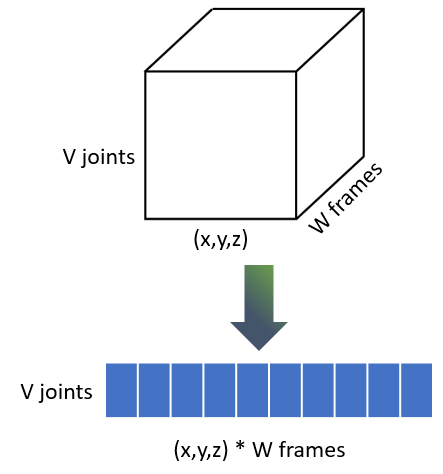}
  \caption{Data pre-processing for GCN architecture.}
  \label{fig:data-preprocess}    
  \end{subfigure}
  \caption{architecture and data processing}
\end{figure}

We use GCN implemented by the DGL python library \cite{wang2019deep-dgl} and ST-GCN implemented as in CTR-GCN\cite{chen2021channel-ctr} as the backbone GNN for our benchmark. For GCN, we use two graph convolutional layers, followed by a sum and max readout as feature extraction layers, and a mlp predictor as classification layer. For ST-GCN, two ST-GCN layers are used.
All layers have 64 hidden units. The GCN architecture is visualized in \cref{fig:gcn}. 

Our benchmark added the data loader for `N-UCLA' and `NTU-RGB+D' skeleton-based action recognition. We adapted the base code from the data loaders of CTR-GCN \cite{chen2021channel-ctr}. We changed the loader to create a DGL compatible graph. The features of each node are preprocessed as shown in \cref{fig:data-preprocess}. As GCN implemented by DGL only accepts 1-D node feature, we concatenated the spatial coordinate information with the temporal information of each joint to create a long vector as the final node feature.

For the data of the `N-UCLA' dataset, we used the prepared data from CTR-GCN \cite{chen2021channel-ctr}, for the data of `NTU-RGB+D', we used the prepared data from the `PaddleVideo' GitHub repository\footnote{\url{https://github.com/PaddlePaddle/PaddleVideo/blob/develop/docs/en/dataset/ntu-rgbd.md}}. We used the train data in the x-sub category to construct our class-incremental learning variant of the dataset.

\section*{\uppercase{Appendix C: Experiments}}\label{C}
We executed all experiments 5 times with the `Tesla V100-SXM3-32GB` GPU. Each task is learned for 100 epochs with $0.001$ learning rate and batch size $10000$, this is equivalent to full batch training, as our tasks contain less than 10000 data points. Each CL process contains 5 tasks, where the model sequentially learns them using the CL method. Each CL process takes around 80 minutes for regularization-based methods and \LWF{} in NTU-CIL, and 10 minutes in N-UCLA-CIL. For \GEM{} and \REPLAY{}, it takes around 100 minutes in NTU-CIL and 15 minutes in N-UCLA-CIL. We compute task-level accuracy $a_{k,k}$ by taking the macro-average of the accuracy of the class in the task. The reported metrics are the average of the 5 executions.

\begin{table}[ht]
\begin{center}
    {\small{
    \caption{Parameter for architectural sensitivity experiment.} \label{tab:arch_param}
\begin{tabular}{lc}
\toprule
Experiment & parameter grid \\
\midrule
width    & width: {[}32,64,128,256,512{]}                                  \\ 
depth    & depth: {[}1,2,4,8,16{]}                                  \\
\bottomrule
\end{tabular}
    }}
\end{center}
\end{table}

For the architectural sensitivity experiment, the exact widths and depths used are presented in \cref{tab:arch_param}. In the width experiment, the depth of GCN is set to default, i.e. two GCN layers. We only changed the width of the GCN layers, and not the classification layers, as we are interested in the impact of changes in graph feature extractions on the CGL performance. In the depth experiment, the width of the GCN layers is set to the default 64 hidden units. 

\begin{table}[ht]
\begin{center}
    {\small{
    \caption{Hyperparameter candidates used for grid search. frac\_memories denotes the percentage of each task that GEM/REPLAY stores as buffer.} \label{tab:hyper}
    \begin{adjustbox}{width=\columnwidth, center}
\begin{tabular}{lc}
\toprule
algorithm & hyperparameter grid \\
\midrule
EWC    & memory\_strength: {[}1,100,10000,1000000{]}                                  \\ 
MAS    & memory\_strength: {[}1,100,10000,1000000{]}                                  \\ 
TWP    & \makecell{lambda\_l: {[}100,10000{]}; lambda\_t: {[}100,10000{]};\\ beta: {[}0.01,0.1{]}} \\ 
LwF    & lambda\_dist: {[}0.1,1,10{]}; T: {[}0.2,2,20{]}                              \\ 
GEM    & \makecell{memory\_strength: {[}0.05, 0.5, 5{]}; \\frac\_memories: {[}0.05, 0.1, 0.2{]}}   \\ 
REPLAY & frac\_memories: {[}0.05, 0.1, 0.2{]}                                         \\ 
\bottomrule
\end{tabular}
\end{adjustbox}
    }}
\end{center}
\end{table} 

To ensure hyperparameter fairness, we conducted a grid search for each algorithm on each dataset using the default task and class order. The grid is shown in \cref{tab:hyper}. Compared to CGLB's \cite{ZhangCGLB} hyperparameter, we replaced the memory buffer size from the number of samples to the fraction of task samples. The two datasets that we implemented are of different sizes. It is fairer to assess the result of rehearsal-based methods when we define the fraction of task data samples to be stored. We report the best hyperparameter we have found for GCN in \cref{tab:best_hyper}.
\end{document}